\documentclass[lettersize,journal]{IEEEtran}
\usepackage{amsmath,amsfonts}
\usepackage{algorithmic}
\usepackage{algorithm}
\usepackage{array}
\usepackage[caption=false,font=normalsize,labelfont=sf,textfont=sf]{subfig}
\usepackage{textcomp}
\usepackage{stfloats}
\usepackage{url}
\usepackage{verbatim}
\usepackage{graphicx}
\usepackage{cite}
\usepackage{amssymb}
\usepackage{multirow}
\usepackage{dsfont}
\usepackage{xcolor}
 
\begin{document}

\title{GALA: A GlobAl-LocAl Approach for \\ Multi-Source Active Domain Adaptation}

\author{Juepeng Zheng,~\IEEEmembership{Member,~IEEE,} 
        Peifeng Zhang, 
        Yibin Wen, \\
        Qingmei Li,
        Yang Zhang, 
        Haohuan Fu,~\IEEEmembership{Senior Member,~IEEE,}
\thanks{Manuscript received . This research was supported in part by the National Natural Science Foundation of China (Grant No. T2125006 and  42401415), the Jiangsu Innovation Capacity Building Program (Grant No. BM2022028). \textit{(Corresponding author: Qingmei Li.)}}

\thanks{Juepeng Zheng is with the School of Artificial Intelligence, Sun Yat-Sen University, Zhuhai, China, also with the National Supercomputing Center in Shenzhen, Shenzhen, China (Email: zhengjp8@mail.sysu.edu.cn).}
\thanks{Peifeng Zhang, Yibin Wen and Yang Zhang  are with the School of Artificial Intelligence, Sun Yat-Sen University, Zhuhai, China (Email:\{zhangpf26, wenyb5, zhangy2583\}@mail2.sysu.edu.cn).}
\thanks{Qingmei Li is with the Tsinghua Shenzhen International Graduate School, Tsinghua University, Shenzhen 518071, China (Email: qingmeili@sz.tsinghua.edu.cn).}
\thanks{Haohuan Fu is with the Tsinghua Shenzhen International Graduate School, Tsinghua University, Shenzhen 518071, China, also with the National Supercomputing Center in Shenzhen, Shenzhen, China, also with the Ministry of Education Key Laboratory for Earth System Modeling and the Department of Earth System Science, Tsinghua University, Beijing 100084, China (e-mail: haohuan@tsinghua.edu.cn).}

}

\markboth{Preprint Version}%
{Shell \MakeLowercase{\textit{et al.}}: A Sample Article Using IEEEtran.cls for IEEE Journals}


\maketitle

\begin{abstract}
Domain Adaptation (DA) provides an effective way to tackle target-domain tasks by leveraging knowledge learned from source domains. Recent studies have extended this paradigm to Multi-Source Domain Adaptation (MSDA), which exploits multiple source domains carrying richer and more diverse transferable information. However, a substantial performance gap still remains between adaptation-based methods and fully supervised learning.
In this paper, we explore a more practical and challenging setting, named Multi-Source Active Domain Adaptation (MS-ADA), to further enhance target-domain performance by selectively acquiring annotations from the target domain. The key difficulty of MS-ADA lies in designing selection criteria that can jointly handle inter-class diversity and multi-source domain variation. To address these challenges, we propose a simple yet effective \underline{G}lob\underline{A}l-\underline{L}oc\underline{A}l strategy (GALA), which combines a global $k$-means clustering step for target-domain samples with a cluster-wise local selection criterion, effectively tackling the above two issues in a complementary manner.
Our proposed GALA is plug-and-play and can be seamlessly integrated into existing DA frameworks without introducing any additional trainable parameters. Extensive experiments on three standard DA benchmarks demonstrate that GALA consistently outperforms prior active learning and active DA methods, achieving performance comparable to the fully-supervised upperbound while using only 1\% of the target annotations.
\end{abstract}


\section{Introduction}
\begin{figure}[t]
    \centering  
    \includegraphics[width=0.5\textwidth]{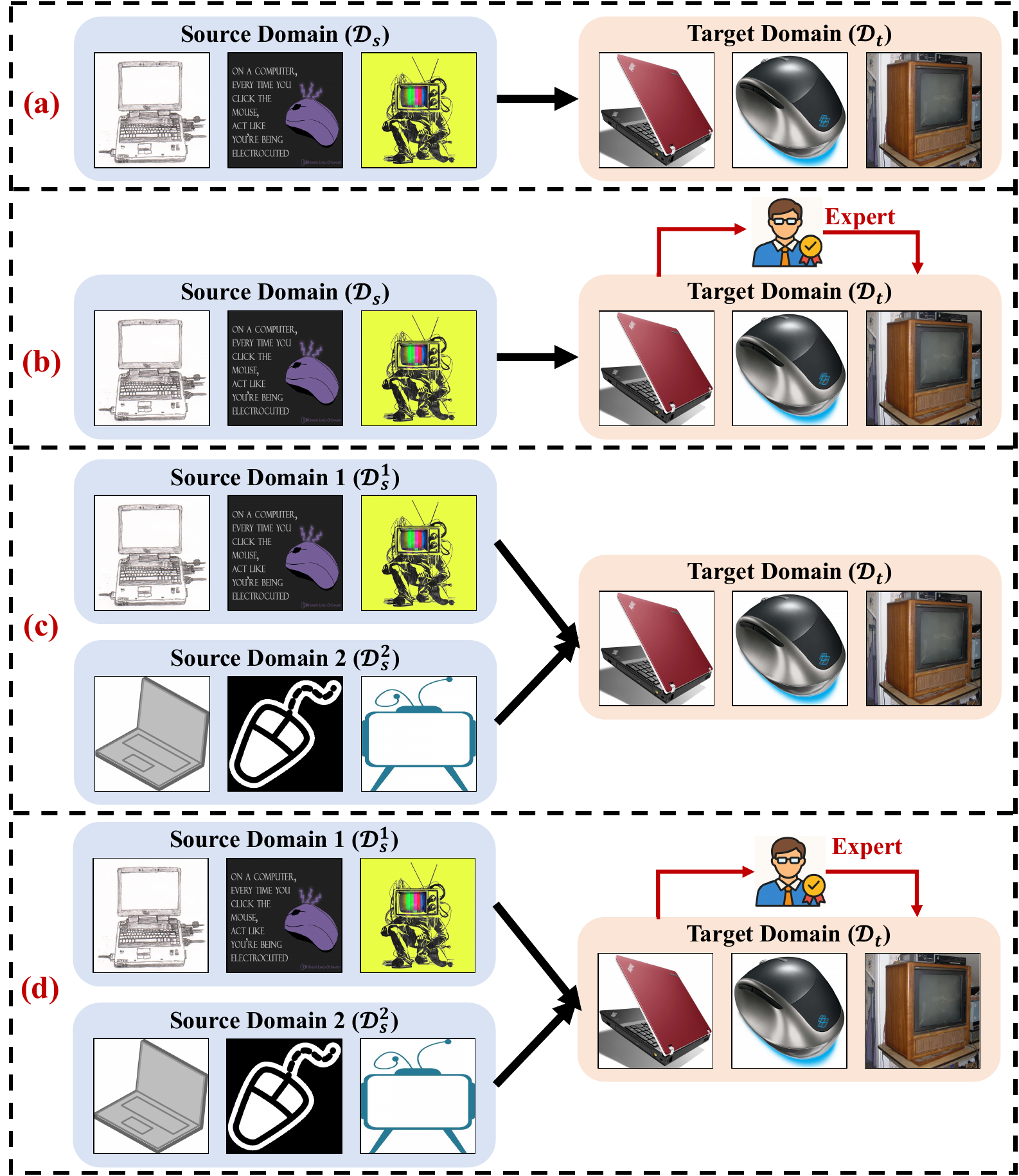}
    \caption{Different DA and ADA settings. (a) Single-Source Domain Adaptation (\textbf{SSDA}), a standard DA setting with only one source domain. (b) Multi-Source Domain Adaptation (\textbf{MSDA}), with multiple source domains from different styles. (c) Single-Source Active Domain Adaptation (\textbf{SS-ADA}), a standard ADA setting by annotating a few images to address domain shifts according to only one source domain. (d) Multiple-Source Active Domain Adaptation (\textbf{MS-ADA}), by annotating a few images to address domain shifts according to multiple source domains with style variations.}
    \label{fig:setting}
\end{figure}

\IEEEPARstart{D}{espite} the great success of deep learning in various fields, it remains a challenge to achieve a generalized deep learning model that can perform well on unseen data, especially given the vast diversity of real-world data and problems. Performance degradation occurs due to various differences between the training data and new test data, for aspects such as statistical distribution, dimension, context, etc. As a result, we usually need to further tune the model for a new target domain. However, with the explosion of data acquisition, additional annotating and tuning soon becomes infeasible. 
Domain adaptation (DA) offers significant benefits in this scenario by adapting the pre-trained models towards new domains with different distributions and properties from the original domain \cite{ben2010theory,yang2021advancing}. Moreover, DA provides great potential for efficiency improvement by reducing the need for retraining on new data from scratch, which has been widely applied in various fields. However, existing DA methods mainly focus on only one source domain and one target domain, \textit{i.e.}, Single Source DA (SSDA) (see Figure \ref{fig:setting}(a)).

Recently, Multi-Source Domain Adaptation (MSDA) (see Figure \ref{fig:setting}(c)) attracts more and more attention since we usually have multiple source domains in practice \cite{ren2023single,li2023enhancing}. In MSDA, labeled samples are given from multiple source domains and we are supposed to make predictions on a target domain, from which only unlabeled samples are available. For instance, images in autonomous driving may be taken under several known weather conditions, or remote sensing data may be collected using different versions of a sensor or a satellite. 
A variety of approaches have been proposed aiming at different application scenarios for MSDA, such as text classification \cite{guo2020multi}, semantic segmentation \cite{zhao2019multi}, person re-identification \cite{ge2019mutual}, etc. Some popular strategies, such as adversarial learning \cite{zhao2018adversarial} and source distilling \cite{zhao2020multi}, have been proposed to improve the performance on target domain using labeled source domains. In all these cases, we wish to learn from all source domains.

\begin{figure*}[t]
    \centering  
    \includegraphics[width=0.95\textwidth]{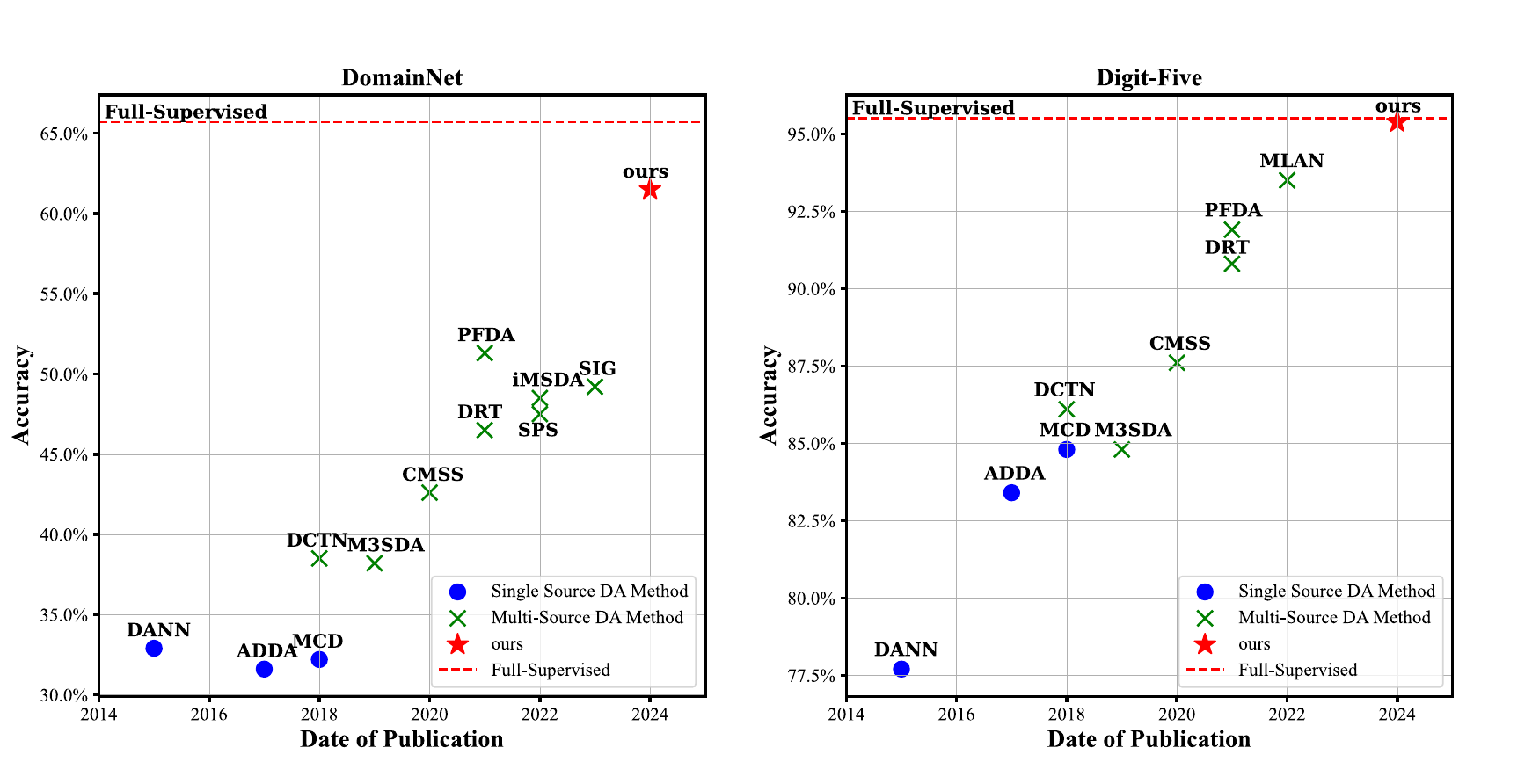}
    \caption{The accuracy of SOTA DA methods on \textbf{Digit-Five} and \textbf{DomainNet} datasets. Although MSDA methods improve the performance compared to SSDA methods, there still exists considerable accuracy gap between the MSDA methods and the Fully-Supervised (FS) upperbound (with 14.4\% and 2.0\% for \textbf{DomainNet} and \textbf{Digit-Five}, respectively). However, our GALA significantly increase the accuracy compared to previous DA methods, and is very close to the FS upperbound with only 1\% annotation of the target dataset.}
    \label{fig:gap}
\end{figure*}

Although abovementioned MSDA approaches could improve the performance of MSDA setting, there still exists considerable accuracy gap between the adaptation  and the Fully-Supervised (FS) upperbound (with 14.4\% and 2.0\% for \textbf{DomainNet} and \textbf{Digit-Five}, respectively) (See Figure \ref{fig:gap}). Therefore, we try to further close the gap between MSDA and FS using active learning for the target domain. Nowadays, Active Domain Adaptation (ADA) methods have been proposed to select unlabeled samples that are uncertain to the model  or representative to the data distribution in the target domain. Most of them focus on Single Source ADA (SS-ADA) (see Figure \ref{fig:setting}(b)) using  density weighted entropy \cite{su2020active},  hard examples mining \cite{xie2022learning} or exploit free energy biases \cite{xie2022active}, etc.

Different from SS-ADA or other active learning settings, we focus on another more practical setting named Multi-Source ADA (MS-ADA) (see Figure \ref{fig:setting}(d)), where we could annotate a few images from the target domain to address domain shifts according to the multiple source domains. 
Although Zhang \textit{et al.} \cite{zhang2024revisiting} firstly propose Detective method for MS-ADA by integrating uncertainty calculation strategy with a dynamic DA framework, they have not considered the specific challenges in MS-ADA.
In fact, there are two major challenges in MS-ADA: \textbf{\textit{Challenge 1}}: the performance imbalance among different classes originates from multiple source domains, so that the active selection algorithm has to ensure inter-class diversity from the perspectives of target dataset;
\textbf{\textit{Challenge 2}}: facing multiple source domains with style and texture variations, the model has to consider a new metric to enhance the transferability of model by conducting suitable target sample selection.
Due to these two challenges, existing ADA methods that directly count on utilizing the uncertainty \cite{su2020active} or free energy \cite{xie2022active} between two specific domains become infeasible.

In this paper, we propose a simple but effective method, named \underline{G}lob\underline{A}l-\underline{L}oc\underline{A}l strategy (GALA), to better tackle the aforementioned two challenges in MS-ADA scenarios. The contributions of this work are as following:

\begin{itemize}
    \item We point out that there are two main challenges in MS-ADA: the inter-class diversity and the multiple source domain variation, and propose a plug-and-play method to any other domain adaptation approaches without adding any trainable parameters.
    \item  We propose a simple yet effective \underline{G}lob\underline{A}l-\underline{L}oc\underline{A}l (GALA) method to overcome abovementioned two challenges. GALA consists of a global $k$-means step for target domain data and a cluster-wise local selection with a new criteria that includes diversity and transferability.
    \item We conduct extensive experiments on three common DA datasets, including \textbf{Digit-Five}, \textbf{Office-31} and \textbf{DomainNet}, and empirical results show that GALA outperforms SOTA ADA methods with considerable gains and achieves comparable results compared with fully-supervised upperbound with only 1\% annotation.
\end{itemize}

Indeed, the global $k$-means step for target domain data mainly addresses the \textbf{\textit{Challenge 1}}, ensuring inter-class diversity from the perspective of target sample selection.
The cluster-wise local selection with a new criteria effectively tackles the \textbf{\textit{Challenge 2}}, reducing the effect of the enriched styles and textures in multiple source domains. To this end, our proposed GALA significantly increase the accuracy compared to previous methods for MS-ADA scenario.

In the remaining sections of this paper, a wide range of related work is introduced in Section \ref{sec:related}. We present a detailed description of our proposed GALA in Section \ref{sec:method}. After that, a thorough experiment with promising results compared with other state-of-the-art approaches is detailed in Section \ref{sec:experiment}. We further discuss our proposed GALA in Section \ref{sec:dis} and summarize it in Section \ref{sec:conc}.

\section{Related Works}
\label{sec:related}

\subsection{Domain Adaptation}

Domain adaptation (DA) aims to minimize domain shift, which refers to the distribution difference between the source and target domains \cite{tian2021heterogeneous,yang2021dual,qin2022deep}.  Most of them focus on Single Source DA (SSDA) (see Figure \ref{fig:setting}(a)). Plenty of single source DA methods have been devised to different tasks including classification \cite{long2016unsupervised,ganin2016domain,qin2022deep}, semantic segmentation \cite{zou2018unsupervised,li2019bidirectional} and object detection \cite{inoue2018cross,hsu2020progressive}. 
To reduce domain discrepancies, there are different kinds of methods. One popular approach is motivated by generative adversarial networks (GAN) \cite{goodfellow2014generative}. Adversarial networks are widely explored and try to minimize the domain shift by maximizing the confusion between source and target, which greatly benefit pixel-level adaptation as well \cite{long2018conditional,zhang2018collaborative,xia2021adaptive}. Alternatively, it is also common practice to adopt statistical measure matching methods to bridge the difference between source and target, including MMD \cite{long2015learning,rozantsev2018beyond}, JMMD \cite{long2017deep}, and other variation metrics \cite{zellinger2016central,kang2019contrastive}.

Although extensive works have been done on SSDA, these methods lead to poor adaptation performance because there is only one single source domain. In this paper, we focus on Multi-Source DA (MSDA), where the model could learn more knowledge from multiple source domains.

\subsection{MSDA}

Compared to the large amount of research done on SSDA, MSDA (see Figure \ref{fig:setting}(c)) is less explored \cite{li2023multidomain}. For example, MDAN \cite{zhao2018adversarial} proposes an adversarial loss to find a representation indistinguishable between all source domains and the target domain. DCTN \cite{xu2018deep} adopts adversarial learning to minimize the discrepancy between the target and each of the multiple source domains.
M$^3$SDA \cite{peng2019moment} minimizes the first order moment-related distance between all source and target domains, which iteratively learns which domains or samples are best suited for aligning to the target.
CMSS \cite{yang2020curriculum} uses an adversarial agent that learns a dynamic curriculum for source samples.
DRT \cite{li2021dynamic} presents dynamic transfer to address domain conflicts, where the model parameters are adapted to samples.
PFDA \cite{fu2021partial} designs three effective feature alignment losses to jointly align the selected features by preserving the domain information of the data sample clusters in the same category and the discrimination between different classes.
SPS \cite{wang2022self} trains several separate domain branch networks with single domains and an ensemble branch network with all domains.
SIG \cite{li2023subspace} incorporates class-aware conditional alignment to accommodate target shifts where label distributions change with the domains.

However, as shown in Figure \ref{fig:gap}, although these above MSDA methods improve the performance compared to SSDA methods, there still exists considerable accuracy gap between the MSDA methods and the Fully-Supervised  upperbound (with 14.4\% and 2.0\% for DomainNet and Digit-Five, respectively). 

\subsection{Active Domain Adaptation}

Traditional active learning methods aim to selectively annotate images by measuring their importance as data uncertainty \cite{gal2017deep,shin2021labor,siddiqui2020viewal,wang2016cost} or diversity \cite{sener2018active,sinha2019variational,chang2021active}. Recently, Active Domain Adaptation (ADA), which instead actively annotates a few images from the target domain to address domain shifts, has received increasing research attention. For example, AADA \cite{su2020active} weighted entropy score with target probability at the equilibrium state of adversarial training. TQS \cite{fu2021transferable} jointly considered committee uncertainty, and domainness for active learning. CLUE \cite{prabhu2021active} integrated entropy score into diversity sampling via $k$-means clustering. SDM \cite{xie2022learning} boosted active domain adaptation accuracy with the help of a margin loss. LAMDA [34] selected domain-representative candidates. However, these methods treated classification and domain alignment as two independent tasks and failed to explicitly consider their correlations. 

However, existing ADA methods focus on Single-Source ADA (SS-ADA) setting (see Figure \ref{fig:setting}(b)). Only \cite{zhang2024d3gu} considers Multi-Target ADA (MT-ADA). Although Zhang \textit{et al.} firstly propose MS-ADA method named Dynamic integrated uncertainty valuation framework (\textbf{Detective}) \cite{zhang2024revisiting}, their model depends on adding more training parameters and is not a universal approach to insert any other MSDA methods. In this paper, we design a novel MS-ADA method named GALA to achieve better target sample selection, which is a simple but effective plug-and-play way for any other domain adaptation methods without adding any trainable parameters, which effectively addresses two major challenges in MS-ADA: inter-class diversity and multiple source domain variations.

\section{Methodology}
\label{sec:method}

\subsection{Problem Setting and Notations}

In contrast to SSDA, MSDA is established on multiple source domains $\mathcal{D}_S=\left\{\mathcal{D}_s^k \right\}_{k=1}^K=\left\{(\mathbf{x}_i^s, \mathbf{y}_i^s)\right\}_{i=1}^{n_S}$, where $K$ is the number of multiple source domains and $n_S=\sum_{k=1}^Kn_s^k$  denotes the total quantity of images in the multi-source domain. The $\mathbf{y}_i^s$ is the corresponding label of $\mathbf{x}_i^s$. 
In the MS-ADA setting, we utilize the labeled multiple source domains and unlabeled target domain $\mathcal{D}_t = \left\{(\mathbf{x}_j^t)\right\}_{j=1}^{n_t}$, where $n_t$ denotes the quantity of images in the target domain,  then selects and annotates images from $\mathcal{D}_t$ at multiple stages to build up selected target set $\mathcal{D}_{st} = \left\{(\mathbf{x}_j^t, \mathbf{y}_j^t)\right\}_{j=1}^{N}$ to join training, in which $N=R \times B$ is the total number of selected target images ($R$ is the total active rounds and $B$ is the quantity of queries for each round).
Hence, if we directly adopt existing SS-ADA algorithms and consider the multiple source domains as one source domain in a brute-force way, the training objective and sample selection strategy will facilitate domain-invariant representations to align the whole source domain $\mathcal{D}_S$ rather than $k$ source domains $\left\{\mathcal{D}_s^k \right\}_{k=1}^K$. 

\begin{figure*}[t]
    \centering  
    \includegraphics[width=1.0\textwidth]{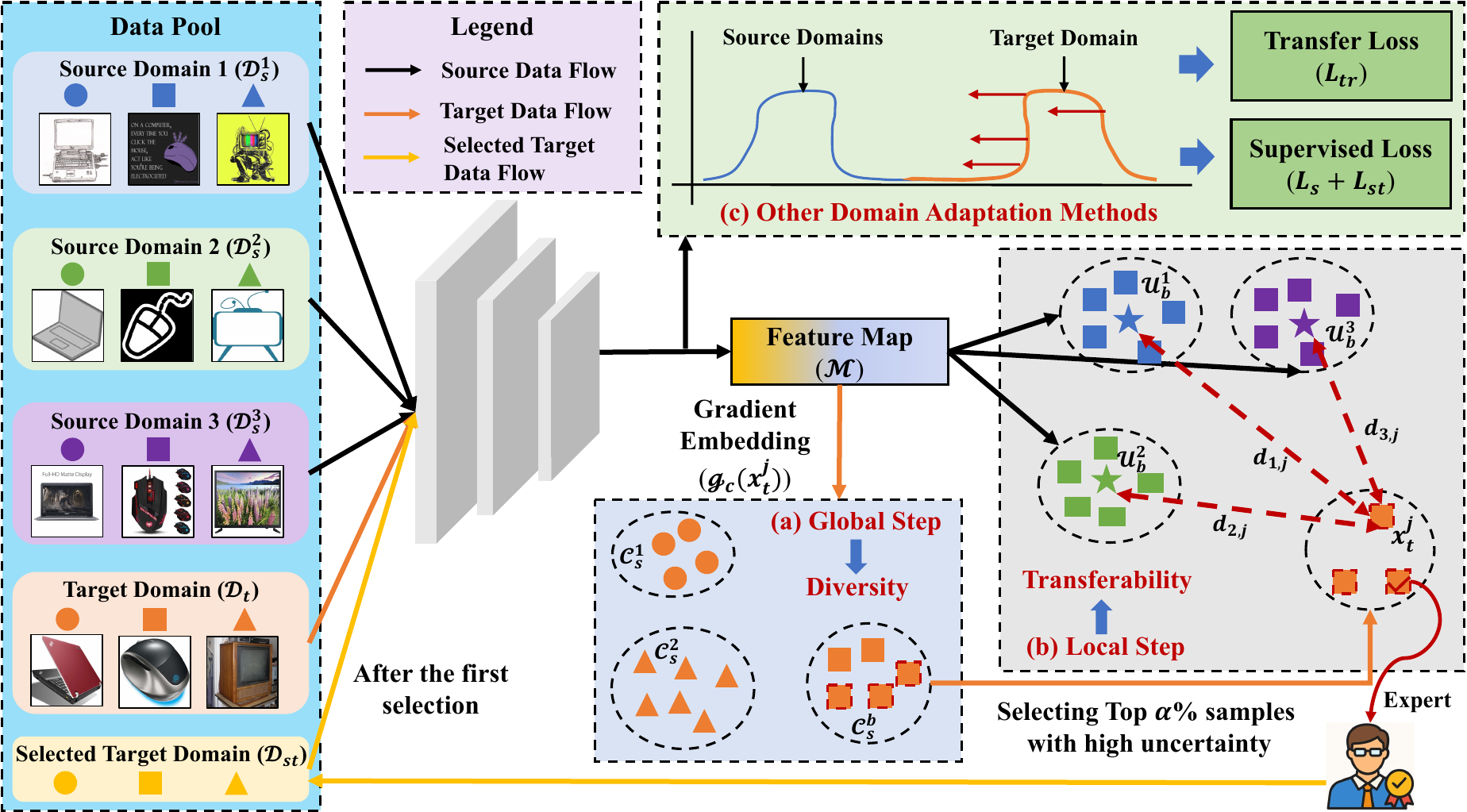}
    \caption{The framework for our proposed GlobAl-LocAl strategy (GALA). (a) \textbf{Global step}: We select the  target queries ($\mathcal{D}_{st}$) to contain both diversity in the gradient embedding space ($g_c\left( x_t^j \right)$) and uncertainty from the perspective of the training model (addressing \textbf{\textit{Challenge 1}}). (b) \textbf{Local step}: We select the highest value with high uncertainty and large domain gap for each cluster ($\mathcal{C}_s^b$) of $v$ according to Eq. \eqref{eq:v} (addressing \textbf{\textit{Challenge 2}}). (c) \textbf{Other DA methods}: Our proposed GALA could be easily plug-and-play to any other DA approaches without any trainable parameters. Pseudo algorithm for GALA is provided in Algorithm \ref{alg2}.}
    \label{fig:framework}
\end{figure*}

As shown in Figure \ref{fig:framework}, our proposed GlobAl-LocAl (GALA) approach can be simply embedded into any DA approaches without adding any trainable parameters, and contains three major modules: the global step, the local step and the specific DA approach. The global step mainly address the \textbf{\textit{Challenge 1}}, ensuring inter-class diversity from the target sample selection 
 perspectives.
The local step effectively tackles the \textbf{\textit{Challenge 2}}, reducing the effect of the enriched styles and textures in multiple source domains.


To better illustrate the workflow and detailed mechanism of our proposed GALA, let us consider a scenario where, at the $r$-th active learning round, the model queries a batch of $B$ unlabeled instances from the target domain. These queried samples are selected according to GALA selection strategy, which balances diversity and transferability. After obtaining the queried samples, the model updates its parameters by incorporating both the newly annotated instances and previously labeled multi-source data, thereby enhancing domain adaptation and alignment.
The entire process is conducted for a total of $R$ active learning rounds, where each round progressively refines the target-domain decision boundary and reduces cross-domain discrepancies. The overall procedure of GALA is summarized in Algorithm \ref{alg2}.

\subsection{Global Step}
\label{sec:global}

Specifically, we utilize $k$-means to achieve the global step on the hypothetical gradient embedding for the target data $\mathcal{D}_t$. Gradient embedding is widely used in conventional active learning algorithm \cite{ash2019deep,venkatesh2020ask}. It is intuitive that the larger the gradient embedding is, the more uncertainty the corresponding sample is. We assume that $z$ is the embedding vector before forwarding to the last layer $W$ for $x_t^j \in \mathcal{D}_{t}^{r-1}$, where $\mathcal{D}_{t}^{r-1}$ denotes the unlabeled target dataset before the annotating round of $r$. As we are not access to the ground-truth labels for the target data, we use their pseudo labels generated from the model. Here, we introduce the gradient of negative cross-entropy loss with respect to the last layer of encoder, which is induced by a pseudo label. Note that we only consider the gradients corresponding to the pseudo label (i.e., $g_{\hat{y}}\left(x_t^j\right)$) for computation efficiency.

\begin{equation}
    g_c \left(x_t^j\right) = - \frac{\partial}{\partial W_c} \mathcal{L}_{CE}\left( x_t^j, \hat{y}; \Theta^r \right) = \mathcal{M} \cdot \left( \mathds{1}_{\left[\hat{y}=c\right]} - p_c \right)
    \label{eq:gradient}
\end{equation}

in which $W_c$ is the weights connected to $c$-th neuron of logits, $\Theta^r$ is the model parameter at the round of $r$ and $\hat{y}$ is the probability corresponding to the pseudo label, which could be formulated as $\hat{y}=\arg \max_{c \in [C]} p_c$.
After that, we calculate $B$ number of centroids ($\mathcal{C}_s^1, ..., \mathcal{C}_s^B$) on the hallucinated gradient space via EM algorithm of $k$-means clustering \cite{macqueen1967classification} by minimizing:

\begin{equation}
    J = \sum_{j=1}^{n_t} \sum_{b=1}^B w_{j, b} \Vert g_{\hat{y}}\left({x_t^j}\right) - \mathcal{C}_s^b \Vert^2
    \label{eq:minimize}
\end{equation}

in which $w_{j,b}$ is an indicator function. If $g_{\hat{y}}\left({x_t^j}\right)$ is assigned to $\mathcal{C}_s^b$, $w_{j,b}=1$. Otherwise $w_{j,b}=0$. Actually, Eq. \eqref{eq:gradient} indicates that the gradient embedding ($g_c\left({x_t^j}\right)$) is a scaling of feature map $\mathcal{M}$, especially with the scale of uncertainty (or confidence). In other words, if a target sample is uncertain to predict (low value of $p_{\hat{y}}$, which also means low confidence), its gradient will be much highly scaled. To this end, Eq. \eqref{eq:minimize} could be considered as a weighted k-means clustering on the feature map through utilizing the function of softmax response \cite{geifman2017selective}. After that, we divide the target samples into $B$ clusters according to their  diversity and uncertainty. Also, we  select a set of $\alpha\%$ candidates with the highest uncertainty calculated by Eq. \eqref{eq:gradient} for each cluster $\mathcal{C}_s^b$.  Therefore, the global step of GALA enables the selected target set ($\mathcal{D}_{st}$) to contain both diversity in the feature embedding space and uncertainty from the perspective of the training model, which effectively addresses \textbf{\textit{Challenge 1}}.

\subsection{Local Step}
\label{sec:local}

After the global step, we divide the target samples into $B$ clusters according to their  diversity and uncertainty. Firstly, we assign all source data from multiple source domains to these $B$ clusters using their feature space embedding ($\mathcal{M}$). After that, we calculate the distance ($d_{k, j}, k=1, ..., K$) of feature embedding between each target data ($x_t^j$) and  these $k$ centroids from different source domains.

\begin{equation}
\label{eq:dis}
    d_{k, j} = \left | \frac{\mu_{s}^{k}}{\sqrt{\left(\sigma_{s}^{k}\right )^2 + \epsilon}} -  \frac{\mu_{t}^{j}}{\sqrt{\left(\sigma_{t}^{j}\right )^2 + \epsilon}}\right |
\end{equation}

where $\epsilon$ is a small number to prevent from numerical problems in case of zero variance. 
$\mu_{s}^{k}$ and $\sigma_{s}^{k}$ are the mean and the standard deviation for the centroid of $k$-th source domain ($\mathcal{U}_b^k$). $\mu_{t}^{j}$ and $\sigma_{t}^{\left(j\right)}$ are the mean and the standard deviation for the centroid of $j$-th target unlabeled data. Therefore, we adopt the max value from different $j$ ( $d_j = \min d_{k, j}, k=1, 2, ..., K$) to describe the distance between the target sample and the multiple source domains. In Sec. \ref{sec:min} we will compare the performance for different ways including the average and the minimum.

To this end, we design a new metric $v$ that can be formulated as Eq. \eqref{eq:v} for the target data $x_t^j$. $v\left( x_t^j \right)$ considers both the domain gap (${d_j}/{\max d}$) generated from Eq. \eqref{eq:dis} using the feature embedding space  and the uncertainty value ($\|g_c\left(x_t^j \right)\|_2$) generated from Eq. \eqref{eq:gradient} using gradient embedding space.  Notably, we normalize the distance ($d_j$) so that it could be $0\sim1$ and calculate the L2-normalization of $g_c\left(x_j \right)$ as the uncertainty.

\begin{equation}
    v\left( x_t^j \right) = \|g_c\left( x_t^j \right)\|_2 \times \frac{d_{j}}{\max d}
    \label{eq:v}
\end{equation}

Therefore we select the target sample with the highest value of $v\left( x_t^j \right)$ for each cluster ($\mathcal{C}_s^b$) that introduced in the global step. This way could effectively pay more attention on those samples that are not very transferable during the sample selection process, which could better address \textbf{\textit{Challenge 2}}.
Finally, we will select $B$ instance at the round of $r$ for annotating. Algorithm \ref{alg2} introduces the overall framework of our proposed GALA. 




\renewcommand{\algorithmicrequire}{\textbf{Input:}} 
\renewcommand{\algorithmicensure}{\textbf{Output:}}

\begin{algorithm}[t]
\caption{MS-ADA framework with GALA algorithm}
\label{alg2}
\begin{algorithmic}[1]
\REQUIRE Initial parameter $\Theta$. Multiple labeled source dataset $\left\{ D_s^k \right\}_{k=1}^K$ Unlabeled target dataset $\mathcal{D}_t^0$. Selected labeled target dataset $\mathcal{D}_{st}^0$. Labeling Budget $B$. Active learning round $R$. 

\ENSURE Well-trained parameter $\Theta^R$. 

\STATE Train a model from the scratch through any DA method using $\left\{ D_s^k \right\}_{k=1}^K$ and $\mathcal{D}_t^0$
\textcolor{blue}{$\#$Global Step}
\FOR{$r = 1:R$}
\STATE For each $x_t^j \in \mathcal{D}_t^{r-1}$, calculate the gradient embedding $g_{\hat{y}}\left( x_t^j \right)$ by Eq. \eqref{eq:gradient}.
\STATE Cluster $\mathcal{D}_t^{r-1}$ into $B$ clusters ($\mathcal{C}_s^1, ..., \mathcal{C}_s^B$) by Eq. \eqref{eq:minimize}.
\STATE Select $\alpha\%$ for $\mathcal{D}_t^{r-1}$ in each $\mathcal{C}_s^b$ cluster with high normalization of $g_{\hat{y}}\left( x_t^j \right)$: $\mathcal{D}_{t, b}^{r-1}$.

\STATE Assign each $x_s^i \in \left\{ D_s^i \right\}_{i=1}^{n_S}$ into these $B$ clusters according to their feature embedding space ($\mathcal{M}_s^i$).
\STATE Selected target sample at round $r$: $\mathcal{X}_t^r = \emptyset$. \\
\textcolor{blue}{$\#$Local Step}
\FOR{$b=1:B$}
\STATE Cluster $\left\{ D_s^i \right\}_{i=1}^k$ that belong to $\mathcal{C}_s^b$  according to the domain label with $k$ centroids ($\mathcal{U}_b^1, ..., \mathcal{U}_b^K$).
\STATE For each $x_t^j \in \mathcal{D}_{t, b}^{r-1}$, calculate the distance between $x_t^j$ and the centroid $\mathcal{U}_b^k$ by Eq. \eqref{eq:dis}.
\STATE Calculate the metric $v \left( x_t^j \right)$ according to Eq. \eqref{eq:v}.
\STATE Select the sample $x_{t, b}^r$ with highest value of $v$ among $\mathcal{D}_{t, b}^{r-1}$ in  each $\mathcal{C}_s^b$.
\STATE $\mathcal{X}_t^r = \mathcal{X}_t^r \cup x_{t, b}^r$.
\ENDFOR
\STATE $\mathcal{D}_t^r = \mathcal{D}_t^{r-1} /\ \mathcal{X}_t^r$
\STATE $\mathcal{D}_{st}^r = \mathcal{D}_{st}^{r-1} \cup \mathcal{X}_{t}^r$

\ENDFOR

\STATE Continue training a model from $\Theta^R$ through any DA method using $\left\{ D_s^k \right\}_{k=1}^K$,  $\mathcal{D}_t^R$ and $\mathcal{D}_{st}^R$.

\RETURN $\Theta^{R*} = \Theta^R$.

\end{algorithmic}
\end{algorithm}

\subsection{Embedding GALA into other DA methods}


In this paper, we integrate our proposed GALA into several well-established domain adaptation (DA) methods to demonstrate its generality and effectiveness. Specifically, we apply GALA to Domain Adversarial Neural Network (DANN) \cite{ganin2016domain}, Moment Matching for Multi-Source Domain Adaptation (M$^3$SDA) \cite{peng2019moment}, Subspace Identification Guarantee (SIG) \cite{li2023subspace}, and Cycle Self-Refinement (CSR) \cite{zhou2024cycle}. In addition, we also directly embed GALA into a baseline CNN model to evaluate its behavior under a standard, non-adversarial DA setting.
For all these variants, we maintain the same supervised loss for the source domain ($L_s$) and the transfer loss ($L_{tr}$) as used in their original implementations, ensuring a fair comparison. When target-domain annotations become available through the active querying process, we further introduce a supervised loss term on the selected target samples ($L_{st}$), which guides the model toward more discriminative and domain-invariant representations.

It is worth emphasizing that GALA can be seamlessly incorporated into a wide range of DA frameworks without introducing any additional trainable parameters. Despite its lightweight nature, it consistently improves target-domain classification accuracy across different architectures and adaptation paradigms, demonstrating its strong generalizability and efficiency compared to existing other ADA approaches.


\begin{figure*}[t]
    \centering  
    \includegraphics[width=1.0\textwidth]{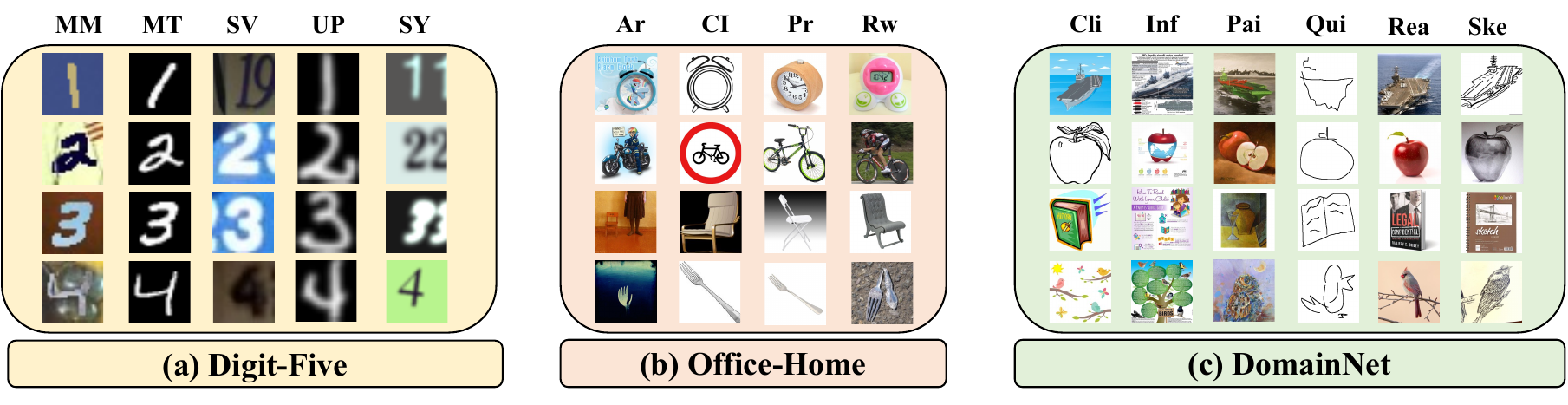}
    \caption{Examples for each dataset: (a) Digit-Five: Images from MNIST-M (MM), MNIST (MT), SVHN (SV), USPS (UP), and Synthetic Digits (SY). (b) Office-Home: Sample images from Artistic (Ar), Clip-Art (Cl), Product (Pr), and Real-World (Rw). (c) DomainNet: Visuals from Clipart (Cli), Infograph (Inf), Painting (Pai), Quickdraw (Qui), Real (Rea), and Sketch (Ske).}
    \label{fig:datasets}
\end{figure*}

\begin{table*}[t]

\centering
\caption{Overall Accuracy (\%) on \textbf{Digit-Five} for state-of-the-art methods. All methods use the LeNet as the backbone  and we set the active ratio as 1\%.}
\begin{tabular}{ccccccccc}
\hline

\textbf{{Method}} &  \textbf{{Active Learning Strategy}} & \textbf{AL or ADA} & \textbf{MM}    & \textbf{MT}  &  \textbf{UP} & \textbf{SV} & \textbf{SY}  &  \textbf{Avg}  \\ \hline
\multirow{10}*{LeNet \cite{lecun1998gradient}} & None & -- & 63.4	& 90.5	& 88.7 & 63.5 & 82.4 & 77.7  \\ 
& Random & AL & 82.0 & 95.9 & 97.0 & 72.0 & 86.5 & 86.7\\
& BADGE [ICLR 2020] \cite{ash2019deep} & AL & 82.3 & 96.2 & 96.8 & 74.9 & 85.4 & 87.1 \\
& APFWA [CVPR 2023] \cite{tejero2023full} & AL & 83.3  & 97.1  & 96.5 &74.3 &84.9 & 87.2 \\
& AADA [WACV 2020] \cite{su2020active} & ADA & 84.5 & 97.3 & 97.9 & 78.5 & 87.0 & 89.0 \\
& CLUE [ICCV 2021] \cite{prabhu2021active} & ADA & 82.7 & 96.7 & 96.3 & 74.6 & 87.2 & 87.5 \\
& SDM [CVPR 2022] \cite{xie2022learning} & ADA & 86.3 & 99.1 & 98.5 & 83.1 & 90.2 & 91.4\\ 
& DUC [ICLR 2023] \cite{xie2023dirichlet} & ADA &  85.3 & 98.6  & 98.1 & 79.6 & 88.2& 90.0 \\
& D$^3$GU [WACV 2024] \cite{zhang2024d3gu} & ADA & 86.8  & 98.1  & 97.2 &80.9 &89.6 & 90.5 \\
& GALA (Ours) & ADA & \textbf{87.6} & \textbf{99.3} & \textbf{99.0} & \textbf{85.6} & \textbf{95.1} & \textbf{93.3} \\\hline
\multirow{10}*{DANN [JMLR 2016] \cite{ganin2016domain}} & None & -- & 70.8	& 97.9	& 93.5 & 67.8 & 86.9 & 83.4  \\ 
& Random & AL & 78.6 & 97.1 & 96.1 & 73.6 & 84.4 & 86.0 \\
& BADGE [ICLR 2020] \cite{ash2019deep} & AL & 80.0 & 96.8 & 96.9 & 74.5 & 86.8 & 87.0 \\
& APFWA [CVPR 2023] \cite{tejero2023full} & AL & 79.6  & 98.0  & 95.6  & 74.7& 85.1& 86.6 \\
& AADA [WACV 2020] \cite{su2020active} & ADA & 83.2 & 97.9 & 97.2 & 78.2 & 87.1 & 88.7 \\
& CLUE [ICCV 2021] \cite{prabhu2021active} & ADA &  80.6 & 94.1 & 96.6 & 69.3 & 85.9 & 85.3 \\
& SDM [CVPR 2022] \cite{xie2022learning} & ADA & 89.0 & 98.9 & 98.7 & 84.2 & 93.3 & 92.8 \\ 
& DUC [ICLR 2023] \cite{xie2023dirichlet} & ADA & 86.6  & 96.9  & 97.0 & 86.1& 87.8& 90.9 \\
& D$^3$GU [WACV 2024] \cite{zhang2024d3gu} & ADA & 87.4  & 96.7  & 97.3  & 83.5 & 89.6 & 90.9 \\
& GALA (Ours) &   ADA & \textbf{91.9} & \textbf{99.2} & \textbf{99.1} & \textbf{87.0} & \textbf{95.3} & \textbf{94.5} \\\hline
\multirow{10}*{M$^3$SDA [ICCV 2019] \cite{peng2019moment}} & None & -- & 72.8 & 98.4 & 96.1 & 81.3 & 89.6 & 87.6 \\ 
& Random & AL & 80.9 & 98.6 & 96.5 & 82.5 & 89.9 & 89.7 \\
& BADGE [ICLR 2020] \cite{ash2019deep} & AL & 82.6 & 98.8 & 97.5 & 82.5 & 92.1 & 90.7  \\
& APFWA [CVPR 2023] \cite{tejero2023full} & AL &  83.1 & 98.6   & 97.3  & 83.1 & 91.7 & 90.8 \\
& AADA [WACV 2020] \cite{su2020active} & ADA & 84.6 & 98.6 & 97.1 & 82.1 & 90.9 & 90.7 \\
& CLUE [ICCV 2021] \cite{prabhu2021active} & ADA & 82.0 & 98.2 & 96.5 & 83.7 & 94.3 & 90.9 \\
& SDM [CVPR 2022] \cite{xie2022learning} & ADA & 91.8 & 98.7 & 98.9 & 86.7 & 97.0 & 94.6 \\ 
& DUC [ICLR 2023] \cite{xie2023dirichlet} & ADA & 88.6  & 98.4   & 97.4  & 84.3 & 94.2 & 92.6 \\
& D$^3$GU [WACV 2024] \cite{zhang2024d3gu} & ADA &  89.4 & 98.2  & 97.8  & 85.6 & 95.1 & 93.2 \\
& GALA (Ours) & ADA  & \textbf{93.3} & \textbf{99.3} & \textbf{99.2} & \textbf{87.9} & \textbf{97.2} & \textbf{95.4} \\\hline
\multicolumn{3}{c}{\textbf{Full-Supervised}} & 94.7 & 99.0 & 99.2 & 87.6 & 97.0 & 95.5  \\ \hline

\end{tabular}
\label{tab:digit}
\end{table*}

\section{Experiments}
\label{sec:experiment}

\subsection{Datasets}


We conduct experiments on three standard domain adaptation benchmarks, including \textbf{Digit-Five}, \textbf{Office-Home} and \textbf{DomainNet}. \textbf{Digit-Five} \cite{xu2018deep} consists of 25,000 images for training and 9,000 images for testing in each domain: MNIST-M (\textbf{MM}), MNIST (\textbf{MT}), SVHN (\textbf{SV}), USPS (\textbf{UP}) and Synthetic Digits (\textbf{SY}). \textbf{Office-Home} \cite{venkateswara2017deep} is a more challenging dataset, which consists of
15,500 images in total from 65 categories of everyday objects. There are four significantly different domain: Artistic images (\textbf{Ar}), Clip-Art images (\textbf{Cl}), Product images (\textbf{Pr}), and Real-World images (\textbf{Rw}).  \textbf{DomainNet} \cite{peng2019moment} is the most challenging and very large scale DA benchmark, which has six different domains: Clipart (\textbf{Cli}), Infograph (\textbf{Inf}), Painting (\textbf{Pai}), Quickdraw (\textbf{Qui}), Real (\textbf{Rea}) and Sketch (\textbf{Ske}). It has around 0.6 million images, including both train and test images, and has 345 different object categories.

\begin{table}[t]
\centering
\caption{Overall Accuracy (\%) on \textbf{Office-Home} for state-of-the-art methods. All methods use the ResNet-50 as the backbone and we set the active ratio as 1\%.  
}
\scriptsize
\begin{tabular}{cccccccc}
\hline

\textbf{{Method}} &  \textbf{{Strategy}} & \textbf{AL/ADA} &  \textbf{Ar} & \textbf{Cl} & \textbf{Pr} & \textbf{Rw} & \textbf{Avg}  \\ \hline
 & None & -- & 64.6	& 52.3	& 77.6 & 80.7 & 68.8  \\ 
& Random & AL &  71.1 & 59.0 & 82.5 & 82.6 & 73.8  \\
{\tiny ResNet-50 \cite{he2016deep}} & AADA  \cite{su2020active} & ADA  & 70.2 & 58.9 & 83.0 & 82.7 & 73.7 \\
\text{\tiny [CVPR 2016]}  & SDM  \cite{xie2022learning} & ADA  & 72.5 & 58.6 & 82.9 & 83.7 & 74.4 \\ 
& D$^3$GU  \cite{zhang2024d3gu} & ADA &  70.9   & 57.6  & 81.3 & 81.8 & 72.9 \\
& GALA  & ADA & \textbf{74.9}  & \textbf{62.0} & \textbf{84.9} & \textbf{85.4} & \textbf{76.8} \\\hline
 & None & -- & 64.3	& 58.0	& 76.4 & 78.8 & 69.4  \\ 
& Random & AL  & 69.3 & 60.6 & 79.4 & 80.7 & 72.5 \\
{\tiny DANN  \cite{ganin2016domain}} & AADA \cite{su2020active} & ADA  & 72.1 & 59.0 & \textbf{84.6} & 82.5 & 74.6 \\
\text{\tiny [JMLR 2016]} & SDM \cite{xie2022learning} & ADA  & 72.5 & 63.0 & 80.6 & 81.6 & 74.4 \\ 
& D$^3$GU \cite{zhang2024d3gu} & ADA &  71.5   &  62.4 & 81.4 & 80.7 & 74.0 \\
& GALA & ADA  & \textbf{74.2}  & \textbf{64.5} & 83.6 & \textbf{84.4} & \textbf{76.7} \\\hline
 & None & -- & 66.2 & 58.6 & 79.5 & 81.4 & 71.4 \\ 
& Random & AL  & 70.9 & 63.1 & 83.5 & 85.4 & 75.7  \\
{\tiny M$^3$SDA  \cite{peng2019moment}} & AADA \cite{su2020active} & ADA  & 74.3 & 65.0 & 86.1 & 87.4 & 78.2 \\
\text{\tiny [ICCV 2019]} & SDM  \cite{xie2022learning} & ADA  & 73.5 & 66.2 & 85.9 & 86.1 & 77.9 \\ 
& D$^3$GU  \cite{zhang2024d3gu} & ADA &  73.1   & 65.5  & 86.2 & 85.3 & 77.5 \\
& GALA & ADA   & \textbf{76.8}& \textbf{67.9} & \textbf{87.5} & \textbf{88.3} & \textbf{80.1} \\\hline
 & None & -- & 76.4 & 63.9  &85.4 & 85.8 & 77.8 \\ 
& Random & AL  & 78.2 & 64.8  & 86.1 & 86.6 & 78.9  \\
{\tiny SIG  \cite{li2023subspace}} & AADA  \cite{su2020active} & ADA  & 79.9 & 65.9 & 87.0 & 87.1 & 80.0 \\
\text{\tiny [NeurIPS 2023]} & SDM  \cite{xie2022learning} & ADA  & 81.5 & 66.7 & 87.9 & 88.5 & 81.2 \\ 
& D$^3$GU  \cite{zhang2024d3gu} & ADA &  81.1 & 65.7 & 86.6 & 88.2 & 80.4 \\
& GALA & ADA   &     \textbf{84.7}& \textbf{67.2} & \textbf{88.2} & \textbf{89.1} & \textbf{82.3} \\\hline
 & None & -- & 76.7 & 71.4 & 86.8 & 85.5 & 80.1 \\ 
& Random & AL  & 77.3 & 70.6  & 86.9 & 86.9 & 80.4  \\
{\tiny CSR \cite{zhou2024cycle}} & AADA \cite{su2020active} & ADA  & 79.2 & 72.3 & 88.3 & 87.5 & 81.8 \\
\text{\tiny [AAAI 2024]} & SDM  \cite{xie2022learning} & ADA  & 79.5 & 73.4 & 87.8 & 88.2 & 82.2 \\ 
& D$^3$GU \cite{zhang2024d3gu} & ADA & 78.2 & 72.6 & 87.4 & 87.9 & 81.5  \\
& GALA & ADA   &      \textbf{84.2}& \textbf{75.3} & \textbf{88.6} & \textbf{89.6} & \textbf{84.4} \\\hline
\multicolumn{2}{c}{Detective \cite{zhang2024revisiting} [CVPR 2024]} &  ADA&82.6&73.6&85.1&85.7&81.8\\\hline


\end{tabular}
\label{tab:home}
\end{table}

\begin{table*}[t]
\centering
\caption{Overall Accuracy (\%) on \textbf{DomainNet} for state-of-the-art methods. All methods use the ResNet-101 as the backbone  and we set the active ratio as 1\%.  
}
\begin{tabular}{cccccccccc}
\hline

\textbf{{Method}} &  \textbf{{Active Learning Strategy}} & \textbf{AL or ADA} &  \textbf{Cli}    & \textbf{Inf}  &  \textbf{Pai} & \textbf{Qui} & \textbf{Rea}    & \textbf{Ske}  &  \textbf{Avg}  \\ \hline
\multirow{10}*{ResNet-101 [CVPR 2016] \cite{he2016deep}} & None & -- & 47.6	& 13.0	& 38.1 & 13.3 & 51.9 & 33.7 & 32.9 \\ 
& Random& AL  & 61.5 & 22.9 & 52.6 & 19.7 & 59.7 & 36.9 & 42.2\\
& BADGE [ICLR 2020] \cite{ash2019deep} & AL & 61.3 & 22.9 & 51.9 & \textbf{20.0} & 62.4 & 40.2 & 43.1 \\
& APFWA [CVPR 2023] \cite{tejero2023full} & AL &  60.9 & 21.2  & 51.5  & 19.3 & 58.6 & 38.4 & 41.7  \\
& AADA [WACV 2020] \cite{su2020active} & ADA & 60.8 & 20.7 & 49.3 & 15.4 & 61.7 & 42.6 & 41.8 \\
& CLUE [ICCV 2021] \cite{prabhu2021active} & ADA & 61.6 & \textbf{23.1} & 52.5 & 17.9 & 64.0 & 54.4 & 45.6 \\
& SDM [CVPR 2022] \cite{xie2022learning} & ADA & 57.6 & 19.7 & 47.3 & 14.9 & 62.5 & 46.6 & 41.4 \\ 
& DUC [ICLR 2023] \cite{xie2023dirichlet} & ADA & 60.9  & 20.1   & 49.7  & 16.5 & 63.1 & 53.8 & 44.0 \\
& D$^3$GU [WACV 2024] \cite{zhang2024d3gu} & ADA & 59.2  & 19.9  & 48.1  & 16.2 & 62.4 & 47.8 & 42.3 \\
& GALA (Ours) & ADA & \textbf{63.4} & 22.7 & \textbf{53.4} & 18.7 & \textbf{64.8} & \textbf{54.8} & \textbf{46.3} \\\hline
\multirow{10}*{DANN [JMLR 2016] \cite{ganin2016domain}} & None & -- & 45.5	& 13.1	& 37.0 & 13.2 & 48.9 & 31.8 & 31.6 \\ 
& Random& AL  & 61.9 & 24.5 & 53.1 & 21.1 & 66.1 & 53.0 & 46.6 \\
& BADGE [ICLR 2020] \cite{ash2019deep} & AL & 64.3 & 23.9 & 55.8 & 23.4 & 65.3 & 54.6 & 47.9 \\
& APFWA [CVPR 2023] \cite{tejero2023full} & AL & 63.1  & 22.7  & 53.2  & 22.9 & 63.2 &  54.9 & 46.7 \\
& AADA [WACV 2020] \cite{su2020active} & ADA & 65.3 & 23.6 & \textbf{56.1} & 20.9 & 66.2 & 55.6 & 48.0 \\
& CLUE [ICCV 2021] \cite{prabhu2021active} & ADA & 66.1 & 28.4 & 54.7 & 21.8 & 61.5 & 57.2 & 48.3 \\
& SDM [CVPR 2022] \cite{xie2022learning} & ADA & 63.1 & 28.4 & 53.1 & 28.6 & 63.3 & 58.4 & 49.2  \\ 
& DUC [ICLR 2023] \cite{xie2023dirichlet} & ADA &  63.7 & 28.6  & 53.3  & 29.1 & 63.7 & 58.6 & 49.5 \\
& D$^3$GU [WACV 2024] \cite{zhang2024d3gu} & ADA &  64.5  & 27.9  & 52.2 & 28.9 & 64.1 & 58.5 & 49.4  \\
& GALA (Ours) & ADA & \textbf{67.9} & \textbf{29.5} & 52.7 & \textbf{29.4} & \textbf{67.0} & \textbf{59.7} & \textbf{51.0} \\\hline
\multirow{10}*{M$^3$SDA [ICCV 2019] \cite{peng2019moment}} & None& -- & 58.6 & 26.0 & 52.3 & 6.3 & 62.7 & 49.5 & 42.6
 \\ 
& Random& AL  & 65.9 & 29.6 & 60.1 & 26.7 & 72.4 & 60.1 & 52.5 \\
& BADGE [ICLR 2020] \cite{ash2019deep}& AL  & 70.6 & 26.3 & 63.9 & 30.8 & 76.3 & 61.7 & 54.9 \\
& APFWA [CVPR 2023] \cite{tejero2023full} & AL &  70.1 &  27.6  & 63.6  & 31.6 & 76.9 & 59.1 & 54.8 \\
& AADA [WACV 2020] \cite{su2020active} & ADA & 68.2 & 28.9 & 62.9 & 29.6 & 74.3 & 59.7 & 53.9 \\
& CLUE [ICCV 2021] \cite{prabhu2021active} & ADA & 69.0 & 29.1 & 65.4 & 32.6 & 72.9 & 60.2 & 54.9 \\
& SDM [CVPR 2022] \cite{xie2022learning} & ADA & 72.1 & 32.4 & 64.3 & 37.1 & 72.8 & \textbf{63.9} & 57.1 \\ 
& DUC [ICLR 2023] \cite{xie2023dirichlet} & ADA &  71.6 & 34.5  & 65.7 & 40.1 & 73.0 & 62.4 & 57.9  \\
& D$^3$GU [WACV 2024] \cite{zhang2024d3gu} & ADA & 72.5  & 33.1  & 64.0 & 39.5 & 73.6 & 61.2 & 57.3 \\
& GALA (Ours) & ADA & \textbf{73.5} & \textbf{36.2} & \textbf{69.8} & \textbf{48.2} & \textbf{79.6} & 63.2 & \textbf{61.5} \\\hline
\multicolumn{3}{c}{\textbf{Full-Supervised}} & 71.6 & 36.7 & 68.6 & 69.6 & 81.8 & 65.8 & 65.7 \\ \hline
\end{tabular}
\label{tab:domainnet}
\end{table*}

\subsection{Training details}

Our methods were implemented based on the PyTorch \cite{paszke2019pytorch}. We adopt LeNet \cite{lecun1998gradient}, ResNet-50 and ResNet-101 \cite{he2016deep} for \textbf{Digit-Five}, \textbf{Office-Home}, and \textbf{DomainNet}, respectively. Both of them are pretrained on the ImageNet dataset \cite{russakovsky2015imagenet}. Whatever module trained from scratch, its learning rate was set to be 10 times that of the lower layers. We adopt mini-batch stochastic gradient descent (SGD) with momentum of 0.95 using the learning rate and progressive training strategies. We conduct 5 active learning stages with equal annotation budget at the epochs of $\{10, 12, 14, 16, 18\}$. According to our Sec. \ref{sec:alpha}, we set $\alpha\%$ as 60\% in our experiments. 
In addition, as \textbf{Office-Home} does not have the test image, we only provide Full-Supervised upperbound for \textbf{Digit-Five} and \textbf{DomainNet} in this paper.
It should be noted that we set the active ratio as 1\% in all experiments and ablation studies.

\textbf{State-of-the-art.} LeNet \cite{lecun1998gradient} or ResNet \cite{he2016deep} are the baseline backbone without any domain adaptation or active learning tricks. We insert our GALA into DANN \cite{ganin2016domain}, M$^3$SDA \cite{peng2019moment}, Subspace Identification Guarantee (SIG) \cite{li2023subspace} and Cycle Self-Refinement method (CSR), which are the classical SSDA method and the classical MSDA method, respectively. 
\textbf{DANN} \cite{ganin2016domain} is a classical DA method and is the cornerstone scheme that adopts adversarial learning and to fool a domain discriminator in a two-player mechanism.
\textbf{M$^3$SD}A \cite{peng2019moment} is a popular MSDA method that aims to transfer knowledge learned from multiple labeled source domains to an unlabeled target domain by dynamically aligning moments of their feature distributions. 
\textbf{SIG} \cite{li2023subspace} leverages variational inference and guarantees the disentanglement of domain-invariant and domain-specific variables under less restrictive constraints regarding domain numbers and transformation properties.
\textbf{CSR} \cite{zhou2024cycle} progressively attempts to learn the dominant transferable knowledge in each source domain in a cycle manner.

Furthermore, we compare our method with existing three AL methods (\textit{i.e.}, Random, BADGE \cite{ash2019deep},  and APFWA \cite{tejero2023full}), five SS-ADA methods (\textit{i.e.}, AADA \cite{su2020active}, CLUE \cite{prabhu2021active}, SDM \cite{xie2022learning}, DUC \cite{xie2023dirichlet} and D$^3$GU \cite{zhang2024d3gu}) and one MS-ADA method (\textit{i.e.}, Detective \cite{zhang2024revisiting}). 
\textbf{Random} denotes we randomly select samples to annotate in the target with a certain percentage.
\textbf{BADGE} \cite{ash2019deep} incorporates both predictive uncertainty and sample diversity into every selected batch.
\textbf{APFWA} \cite{tejero2023full} incorporates both full and weak supervision, which dynamically determines the proportion of full and weak annotations within each batch.
\textbf{AADA} \cite{su2020active} uses a domain discriminative model to align domains and utilizes the model to weigh samples to account for distribution shifts.
\textbf{CLUE} \cite{prabhu2021active} performs uncertainty-weighted clustering to identify target instances for labeling that are both uncertain under the model and diverse in feature space.
\textbf{SDM} \cite{xie2022learning} consists of a maximum margin loss and a margin sampling algorithm for data selection.
\textbf{DUC} \cite{xie2023dirichlet} simultaneously achieves the mitigation of miscalibration and the selection of informative target samples.
\textbf{D$^3$GU} \cite{zhang2024d3gu} align different domains and actively select samples from them for annotation. As D$^3$GU is designed for multi-target domains, we set the domain number in D$^3$GU as 1 in our experiments.
In addition, we also compare our performance with \textbf{Detective} \cite{zhang2024revisiting} on the Office-Home dataset, as it is the only method that concentrates on the MS-ADA scenario and only publicly provides experimental results on this dataset. \textbf{Detective} comprehensively considers
domain and predictive uncertainty, and context diversity to select informative target samples.
Notably,  we set the active ratio as 1\% in all experiments and ablation studies in this paper.

\subsection{Results}


\textbf{Experiments on Digit-Five.}
As reported in Table \ref{tab:digit}, our method GALA overpasses all other active learning strategies in different domain adaptation scenarios by at least $1.2\%$ improvement on average. Among all methods, SDM achieves the second-highest accuracy across all tasks, performing the strongest as a comparison approach. However, our method still demonstrates superior performance over SDM especially in challenging tasks. For example, with M$^3$SDA method, the accuracy of GALA is $1.5\%$ and $1.2\%$ higher than that of SDM on tasks \textbf{MM} and \textbf{SV}, respectively. GALA approach achieves the state-of-the-art in the MS-ADA setting with M$^3$SDA method and is very close to Fully-Supervised. It is remarkable that GALA achieves higher accuracy on transfer tasks of \textbf{MT}, \textbf{UP} and \textbf{SY} compared to Full-Supervised.

\textbf{Experiments on Office-Home.}
The accuracy on \textbf{Office-Home} are shown in Table \ref{tab:home}. We can observe that our GALA achieves the highest average accuracy ($76.8\%$, $76.7\%$ and $80.1\%$) compared to other active learning strategies in these three methods. Notably,  when benchmarked against \textbf{Detective} which is dedicated to the MS-ADA scenario, GALA still shows clear advantages on this dataset.When datasets become more complex, the effectiveness of most active learning strategies begins to diminish. For example, SDM, which performs second best on the \textbf{Digit-Five} dataset, does not perform as well as AADA in several scenarios. However, our method is able to maintain the efficiency. Especially, GALA promotes the average accuracy by at least $2.2\%$ for all methods. In the most challenging task \textbf{Cl}, GALA is the only strategy with more than $60\%$ accuracy and surpasses $2.1\%$ over the state-of-the-art active learning strategy CLUE in backbone setting. GALA outperforms all existing active learning strategies with M$^3$SDA method except with DANN on Pr and achieves the state-of-the-art on the MS-ADA setting.

\textbf{Experiments on DomainNet.}
Table \ref{tab:domainnet} lists the results on \textbf{DomainNet}, the most difficult dataset to date. Our proposed GALA improves the average accuracy by at least $0.7\%$ (CLUE \cite{prabhu2021active} vs. GALA in ResNet-101) compared to any other active learning strategy with any method listed in this table. When our transfer tasks become complex and diverse, different methods exhibit distinct advantages. Each comparative method, except for random, achieves the highest accuracy in a specific task. However, our method remains superior, achieving optimal performance in 14 out of 18 transfer scenarios with different methods. In addition, it is encouraging that GALA overpasses SDM with $3.8\%$ and $11.1\%$ improvement in two of the most difficult tasks \textbf{Inf} and \textbf{Qui}, respectively. Furthermore, GALA with M$^3$SDA is higher (with $1.9\%$ and $1.2\%$) than Full-supervised in \textbf{Cli} and \textbf{Pai}.

\subsection{Sensitive analysis of $\alpha$} 
\label{sec:alpha}

\begin{figure*}[t]
    \centering  
    \includegraphics[width=0.9\textwidth]{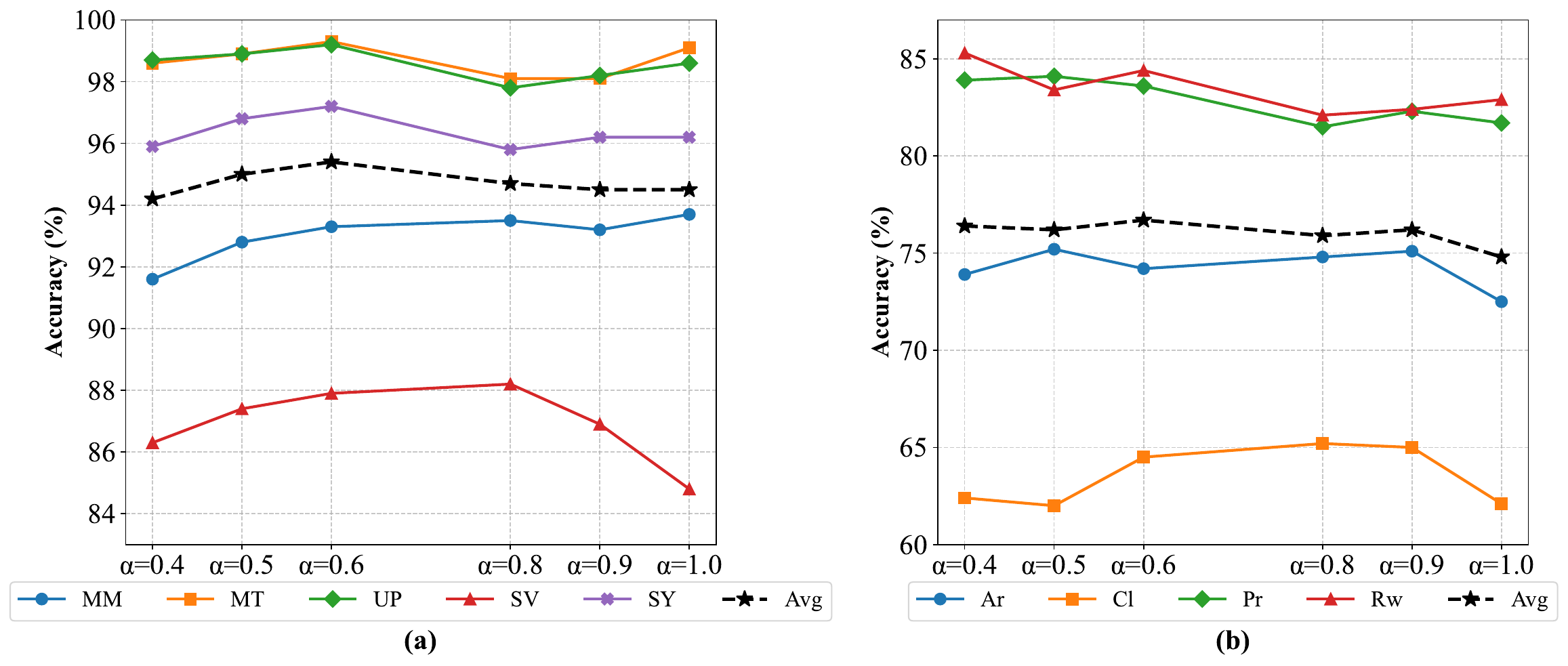}
    \caption{Sensitive analysis of $\alpha$ introduced in Sec. \ref{sec:global}: (a) Performance on different value of $\alpha$ for \textbf{Digit-Five} (b) Performance on different value of $\alpha$ for \textbf{Office-Home}}
    \label{fig:alpha}
\end{figure*}


\begin{table*}[t]
\centering
\caption{Overall Accuracy (\%) on four general DA datasets using different distance metrics.}
\begin{tabular}{ccccccc}
\hline
\textbf{Method} & \textbf{Distance type} &  \textbf{Digit-Five}   & \textbf{Office-Home}  & \textbf{DomainNet} &\textbf{Avg}  \\ \hline

ResNet \cite{he2016deep} & None &  77.7  & 68.8 & 32.9 & 59.8    \\ 
GALA  & $\mu$   &  90.9 & 73.5 & 42.8 & 69.1 \\
GALA & Wasserstein &  92.6 & 74.2 & 46.1 & 71.0  \\

GALA & $\mu / \sqrt{\sigma^2+\epsilon}$  & \textbf{93.3} & \textbf{76.8} & \textbf{47.0} & \textbf{72.4}    \\\hline

DANN \cite{ganin2016domain} & None   & 83.5 & 69.4 & 31.6 & 61.5  \\
DANN + GALA & $\mu$ &   93.1 & 74.4 & 46.8 & 71.4 \\
DANN + GALA & Wasserstein &  94.2 & 75.9 & 50.6 & 73.6 \\
DANN + GALA & $\mu / \sqrt{\sigma^2+\epsilon}$ &  \textbf{94.5} & \textbf{76.7} & \textbf{51.0} & \textbf{74.1} \\\hline

M$^3$SDA \cite{peng2019moment} & None &  87.6 & 71.4 & 42.6 & 67.2 \\ 
M$^3$SDA + GALA & $\mu$ &  94.3 & 77.0 & 59.3 & 76.9   \\
M$^3$SDA + GALA & Wasserstein  &  94.8 & 80.0 & 61.9 & 78.9 \\
M$^3$SDA + GALA & $\mu / \sqrt{\sigma^2+\epsilon}$ & \textbf{95.4} & \textbf{80.1} & \textbf{61.5} & \textbf{79.0}   \\\hline
\multicolumn{2}{c}{Full-Supervised} & 95.5 & None & 65.7 & None \\ \hline
\end{tabular}
\label{office-dis}
\end{table*}

As introduced in Sec. \ref{sec:global}, after the global step, we first select a set of $\alpha\%$ (the only hyper-parameter in GALA.) candidates with the highest uncertainty calculated by Eq. \eqref{eq:gradient} for each cluster $\mathcal{C}_b$. Figure \ref{fig:alpha} (a) and (b) displays the performance under different  $\alpha$ for \textbf{Digit-Five} and \textbf{Office-Home}. We can observe that when $\alpha = 60$, that means we select top $60\%$ target samples with high uncertainty for each cluster ($\mathcal{C}_b$) could achieve higher performance. This strategy filters those target samples with low uncertainty but large domain gap, which actually already have high confidence in label prediction.

\section{Discussions}
\label{sec:dis}

\subsection{Different distance metric} 
\label{sec:abla}
We compare another two distance metrics with $\mu / \sqrt{\sigma^2+\epsilon}$. The one is $\left|\mu_s^i - \mu_t^j \right|$ and the other is Wasserstein distance \cite{villani2009wasserstein}: ${\left \| \mu_s^i - \mu_t^j \right \|}^2_2 + \left(\right(\sigma_s^i\left)^2 + \right(\sigma_t^j\left)^2 - 2\sigma_s^i\sigma_t^j\right)$.
Table \ref{office-dis} qualifies the performance of different distance metrics under the same conditions. We can observe that $\mu / \sqrt{\sigma^2+\epsilon}$ achieves higher overall average accuracy (mean accuracy for four average accuracies of four DA datasets) than other two distance metrics. The reason maybe that the distance calculated only between means $\mu$ can not represent the characteristic of variances $\sigma$, and the Wasserstein distance does not consider the balance for relative impact between means and variances. On the other hand, our distance calculation scheme is consistent with BN \cite{ioffe2015batch} and TransNorm \cite{wang2019normalization}, representing the distance between means with variance normalization and performing the best results among these three distance ways. Additionally, we can observe that GALA has the highest increase without any domain loss module. By embedding GALA to ResNet, M$^3$SDA  has lower improvement room than DANN. The reason maybe that M$^3$SDA  is a more well-rounded DA method for MSDA setting so that is closer the upper bound and has less space to boost.

\begin{table*}[t]
\centering
\caption{Performance comparisons between the feature map ($\mathcal{M}\left( x_t^j \right)$) and the gradient embedding (${g}_c\left( x_t^j \right)$) for \textbf{Digit-Five}. All methods use the LeNet as the backbone.}
\begin{tabular}{ccccccccccc}
\hline
\multirow{2}*{\textbf{Method}} & \multicolumn{2}{c}{\textbf{{Global}}} & \multicolumn{2}{c}{\textbf{{Local}}}  & \multirow{2}*{\textbf{MM}}  & \multirow{2}*{\textbf{MT}}  & \multirow{2}*{\textbf{UP}} & \multirow{2}*{\textbf{SV}} & \multirow{2}*{\textbf{SY}}  & \multirow{2}*{\textbf{Avg}}  \\ 
& {$\mathcal{M}\left( x_t^j \right)$} & ${g}_c\left( x_t^j \right)$ & {$\mathcal{M}\left( x_t^j \right)$} & ${g}_c\left( x_t^j \right)$ \\\hline
\multirow{4}*{LeNet \cite{lecun1998gradient} + GALA} & $\checkmark$ & & $\checkmark$ & & 88.6 & 98.9 & 98.5 & 83.9 & 94.9 & 93.0 \\ 
&$\checkmark$ & & &$\checkmark$ & 89.6 & 98.3 & 97.9 & 84.0 & 92.5 & 92.5 \\
& & $\checkmark$ & $\checkmark$& & 87.6 & 99.3 & 99.0 & 85.6 & 95.1 & 93.3 \\
& & $\checkmark$& &$\checkmark$ & 86.9 & 98.5 & 98.2 & 85.3 & 95.2 & 92.8 \\ \hline
\multirow{4}*{DANN \cite{ganin2016domain} + GALA} & $\checkmark$ & & $\checkmark$  & & 90.0 & 99.0 & 99.0 & 87.1 & 93.3 & 93.7 \\
&$\checkmark$ & & &$\checkmark$ & 90.5 & 98.4 & 98.7 & 86.9 & 94.9 & 93.9 \\
& & $\checkmark$ & $\checkmark$& & 91.9 & 99.2 & 99.1 & 87.0 & 95.3 & 94.5 \\
& & $\checkmark$& &$\checkmark$& 90.1 & 98.9 & 98.4 & 87.2 & 94.8 & 93.9 \\ \hline
\multirow{4}*{M$^3$SDA \cite{peng2019moment} + GALA} & $\checkmark$ & & $\checkmark$& & 92.9 & 99.1 & 98.8 & 84.2 & 96.0 & 94.2  \\ 
&$\checkmark$ & & &$\checkmark$ & 93.0 & 89.7 & 99.0 & 86.3 & 96.8 & 93.0\\
& & $\checkmark$ & $\checkmark$& & 93.3 & 99.3 & 99.2 & 87.9 & 97.2 & 95.4 \\
& & $\checkmark$& &$\checkmark$& 93.5 & 99.0 & 98.4 & 87.1 & 96.3 & 94.9  \\\hline
\end{tabular}
\label{tab:embed_digit}
\end{table*}

\begin{table*}[t]
\centering
\caption{Performance comparisons between the feature map ($\mathcal{M}\left( x_t^j \right)$) and the gradient embedding (${g}_c\left( x_t^j \right)$) for \textbf{Office-Home}. All methods use the ResNet-50 as the backbone.}
\begin{tabular}{cccccccccc}
\hline
\multirow{2}*{\textbf{Method}} & \multicolumn{2}{c}{\textbf{{Global}}} & \multicolumn{2}{c}{\textbf{{Local}}}  & \multirow{2}*{\textbf{Ar}}  & \multirow{2}*{\textbf{Cl}}  & \multirow{2}*{\textbf{Pr}} & \multirow{2}*{\textbf{Rw}}   & \multirow{2}*{\textbf{Avg}}  \\ 
& {$\mathcal{M}\left( x_t^j \right)$} & ${g}_c\left( x_t^j \right)$ & {$\mathcal{M}\left( x_t^j \right)$} & ${g}_c\left( x_t^j \right)$ \\\hline
\multirow{4}*{ResNet-50 \cite{he2016deep} + GALA} & $\checkmark$ & & $\checkmark$ & & 73.2 & 61.8 & 84.2 & 84.5 & 75.9  \\ 
&$\checkmark$ & & &$\checkmark$ & 73.6 & 59.1 & 84.7 & 83.9 & 75.3\\
& & $\checkmark$ & $\checkmark$& & 74.9 & 62.0 & 84.9 & 85.4 & 76.8 \\
& & $\checkmark$& &$\checkmark$& 72.7 & 62.7 & 83.9 & 85.6 & 76.2 \\ \hline
\multirow{4}*{DANN \cite{ganin2016domain} + GALA} & $\checkmark$ & & $\checkmark$  & & 73.9 & 61.8 & 83.5 & 85.3 & 76.1 \\
&$\checkmark$ & & &$\checkmark$ & 74.8 & 65.6 & 83.4 & 84.7 & 77.1 \\
& & $\checkmark$ & $\checkmark$& & 74.2 & 64.5 & 83.6 & 84.4 & 76.7 \\
& & $\checkmark$& &$\checkmark$& 74.0 & 63.9 & 83.2 & 84.1 & 76.7 \\ \hline
\multirow{4}*{M$^3$SDA \cite{peng2019moment} + GALA} & $\checkmark$ & & $\checkmark$& & 75.6 & 65.0 & 87.8 & 85.1 & 78.4  \\ 
&$\checkmark$ & & &$\checkmark$& 74.3 & 66.5 & 86.9 & 87.4 & 78.8 \\
& & $\checkmark$ & $\checkmark$& & 76.8 & 67.9 & 87.5 & 88.3 & 80.1 \\
& & $\checkmark$& &$\checkmark$& 76.2 & 66.3 & 86.1 & 88.6 & 79.3 \\\hline
\end{tabular}
\label{tab:embed_office}
\end{table*}

\begin{figure*}[t]
    \centering  
    \includegraphics[width=0.85\textwidth]{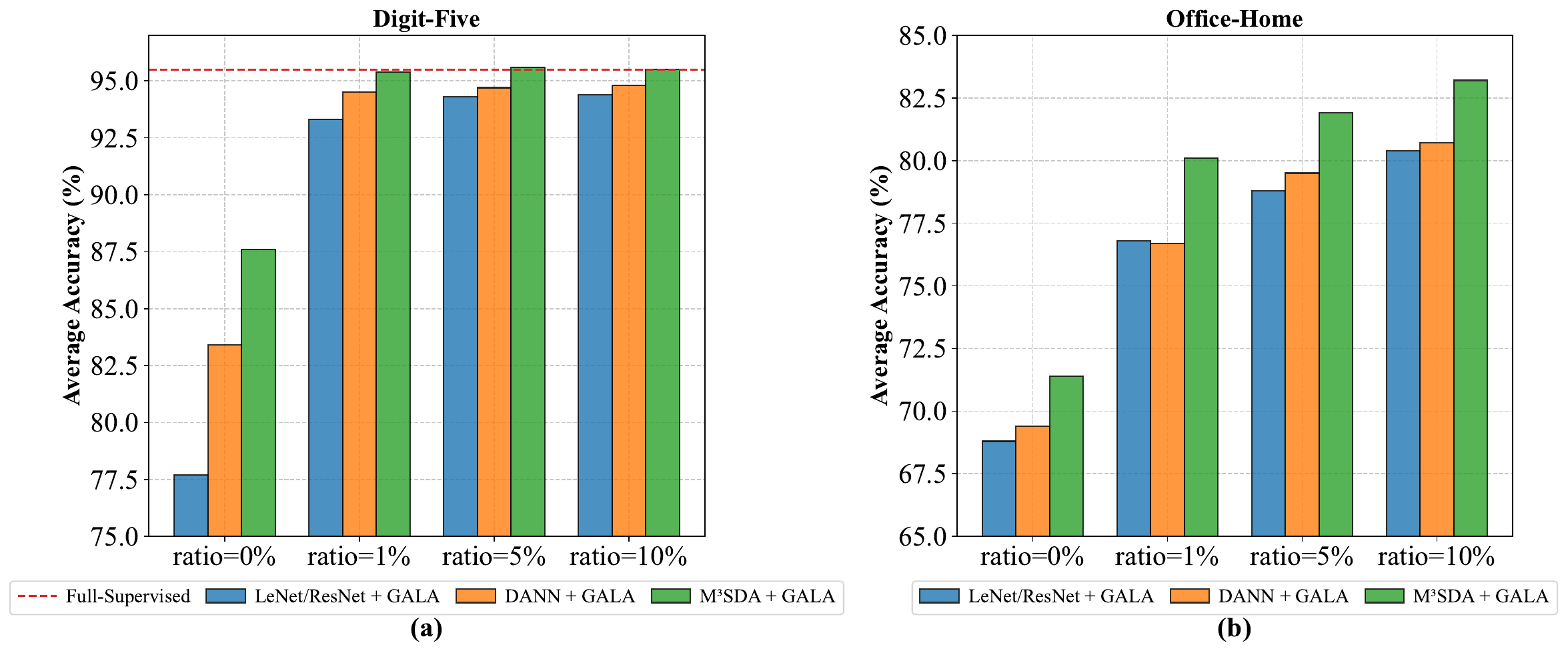}
    \caption{Average accuracy on different ratios including 1\%, 5\%, 10\% for (a) \textbf{Digit-Five} and (b) \textbf{Office-Home}}
    \label{fig:ratio}
\end{figure*}

\subsection{Feature Map ($\mathcal{M}\left( x_t^j \right)$) or Gradient Embedding (${g}_c\left( x_t^j \right)$)?}
\label{sec:embed}

Gradient embedding (${g}_c\left( x_t^j \right)$) is widely used in conventional active learning algorithm \cite{ash2019deep,venkatesh2020ask}, which can be represented as the uncertainty of the sample.
Also, feature map ($\mathcal{M}\left( x_t^j \right)$) could calculate the distance between any two samples or domains \cite{long2015learning,long2018conditional}. Therefore, the gradient embedding and the feature map could both use in the global step and the local step. Table \ref{tab:embed_digit} and Table \ref{tab:embed_office} list the performance comparisons between the feature map ($\mathcal{M}\left( x_t^j \right)$) and the gradient embedding (${g}_c\left( x_t^j \right)$) for \textbf{Digit-Five} and  \textbf{Office-Home}, respectively. Experimental results show that when we choose the gradient embedding for the global step and the feature map for the local step, the final average accuracy performs best.

\begin{figure*}[t]
    \centering  
    \includegraphics[width=0.85\textwidth]{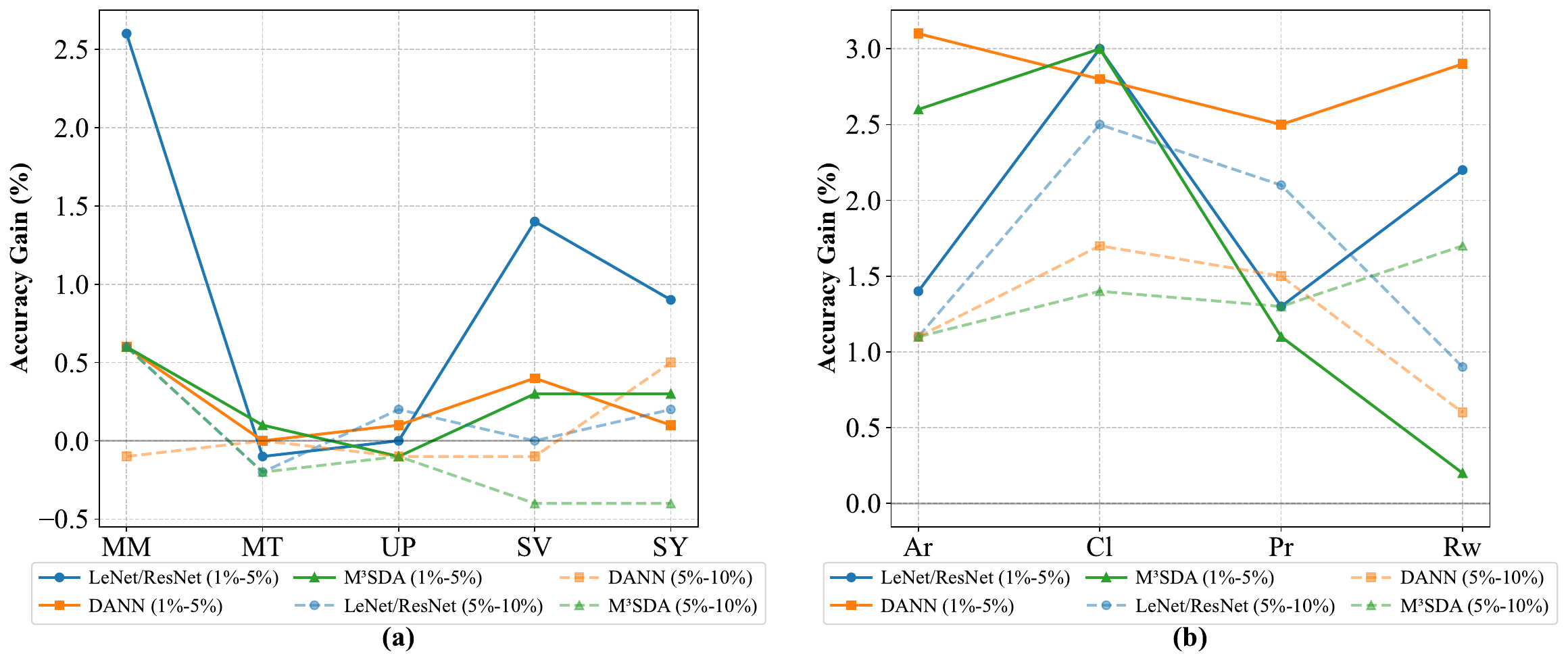}
    \caption{Accuracy gain when increasing active ratio including 1\%-5\% and 5\%-10\% for (a) \textbf{Digit-Five} and (b) \textbf{Office-Home}}
    \label{fig:gain}
\end{figure*}

\subsection{Performance on Different Active Ratios}


Figure \ref{fig:ratio} (a) and (b) display average accuracy for different active ratios including 1\%, 5\%, and 10\% for Digit-Five and Office-Home, respectively. We can observe that the accuracy consistently increases as the active ratio rises, except for M$^3$SDA + GALA for Digit-Five where it shows no significant change or even slightly decreases. The performance of some transfer tasks exceeds fully-supervised learning, such as M$^3$SDA + GALA with just 5\% labeled target data. Additionally, Figure \ref{fig:gain} (a) and (b) illustrate the accuracy gain when increasing the active ratio from 1\% to 5\% and from 5\% to 10\% with different methods across various tasks for Digit-Five and Office-Home, respectively.
We can observe that in most scenarios, increasing the active ratio leads to a corresponding steady improvement in performance. However, in some cases, increasing the annotation ratio actually results in decreased accuracy, for example, with M$^3$SDA in SV and SY tasks. Surprisingly, integrating GALA with other DA methods consistently outperforms other active learning or active DA methods with substantial gains and achieves comparable results to the fully supervised upper bound with only 1\% annotation.




\subsection{Average or Minimum for Calculating $d$?}
\label{sec:min}

\begin{figure}[t]
    \centering  
    \includegraphics[width=0.9\columnwidth]{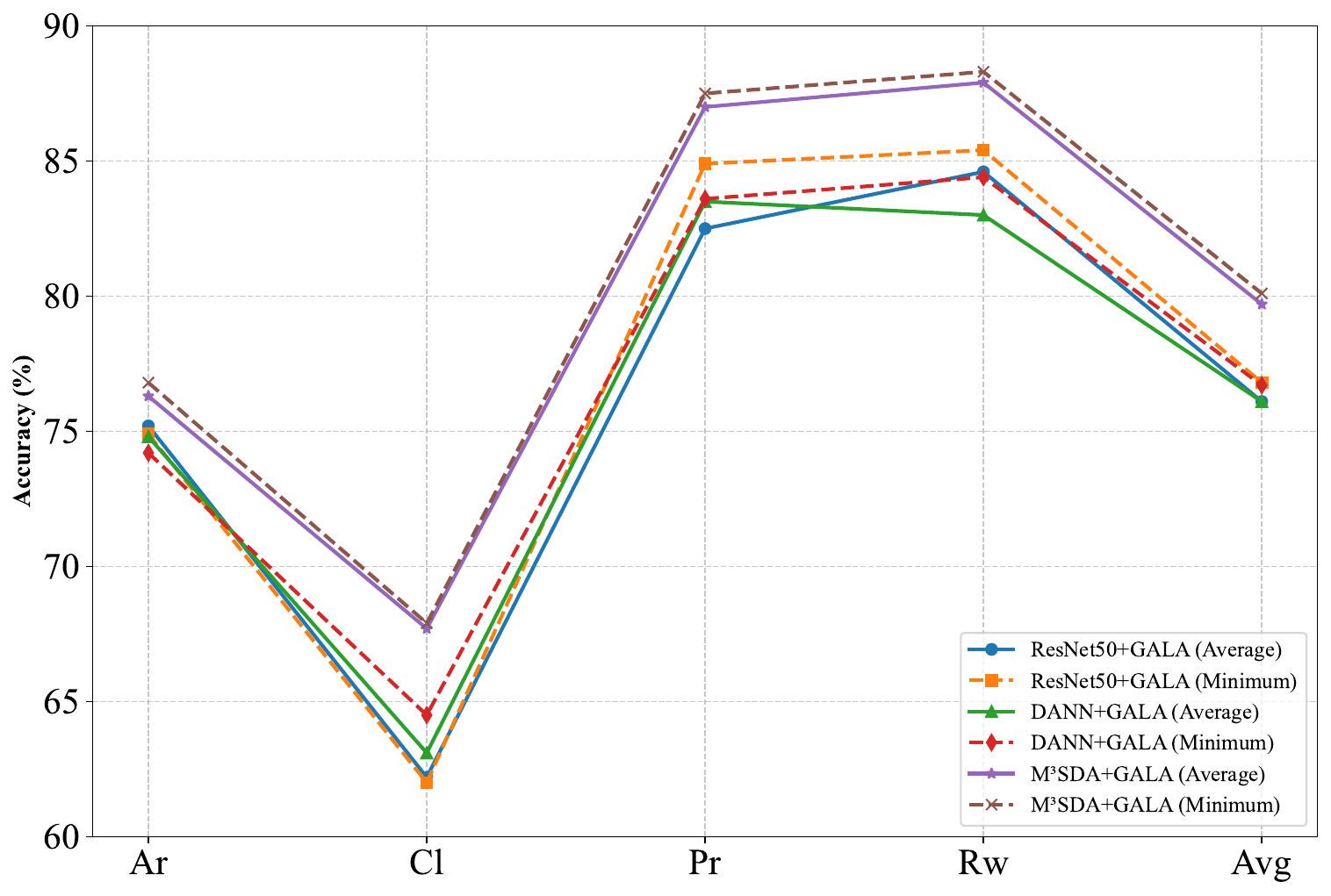}
    \caption{Comparisons in \textbf{Office-Home} between the average and the minimum for $d$ in Eq. \eqref{eq:v}. All methods use the ResNet-50 as the backbone.}
    \label{fig:d}
\end{figure}

Figure \ref{fig:d} shows the performance in \textbf{Office-Home} between the average and the minimum for $d$ in Eq. \eqref{eq:v} (see detail in Sec. \ref{sec:local}). We can observe that the performance of minimum is better than the average way in most cases. We select the minimum distance among $d_j = \min d_{k, j}, k=1, ..., K$. If $d_j$ is large, it means that the target sample ($x_t^j$) with high uncertainty is far away from any source domain, which maybe crucially important for the target domain. However, if we use the average distance of $d_j = 1/K \sum_{k=1}^K d_{k, j}$, although the $d_j$ is large because the target sample is far away from some source domains, once a specific source domain ($\mathcal{D}_s^k$) is small, the target sample ($x_t^j$) could still learn a good representation from the specific source domain. To this end, we select the minimum distance as the domain gap among $d_{k, j}$ for each target sample $x_t^j$.

\subsection{In-depth theoretical analysis}

\begin{figure}[t]
    \centering  
    \includegraphics[width=0.5\textwidth]{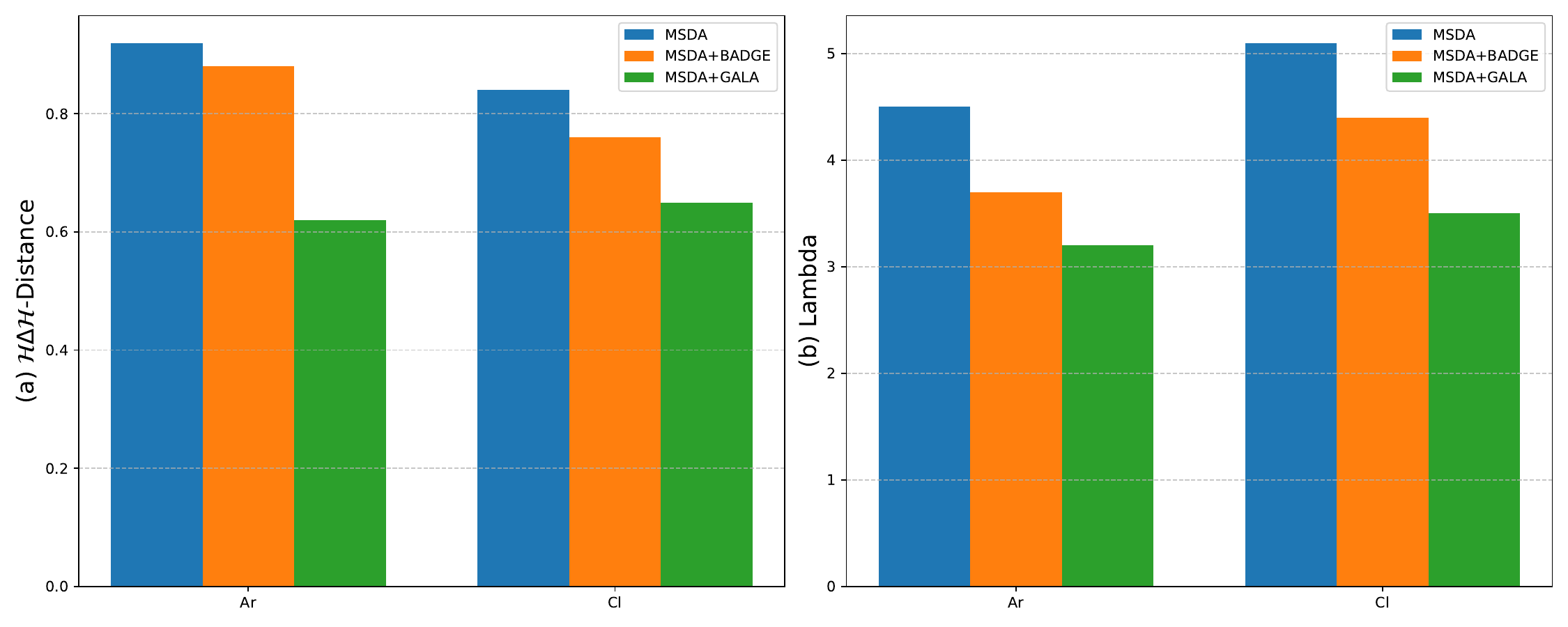}
    \caption{In-depth theoretical analysis of $\mathcal{H} \Delta \mathcal{H}$-distance and $\lambda$.}
    \label{fig:In-depth}
\end{figure}

Following the DA theory, the upper boundary of the expected error $\mathcal{E}_\mathcal{T}\left(h\right)$ of a hypothesis $h$ on the target domain, can be expressed as the sum of three terms ($\mathcal{E}_\mathcal{T}\left(h\right) \leq \mathcal{E}_\mathcal{S} \left(h\right) + \frac{1}{2} d_{\mathcal{H} \Delta \mathcal{H}}\left(\mathcal{S}, \mathcal{T}\right) + \lambda$):
(\textbf{$i$}) expected error of $h$ on the source domain, $\mathcal{E}_\mathcal{S} \left(h\right)$; (\textbf{$ii$}) $\mathcal{H} \Delta \mathcal{H}$-distance $d_{\mathcal{H} \Delta \mathcal{H}}\left(\mathcal{S}, \mathcal{T}\right)$, measuring domain shift as the discrepancy between the disagreement of two hypotheses $h, h' \in \mathcal{H} \Delta \mathcal{H}$; and (\textbf{$iii$}) the error $\lambda$ of the ideal joint hypothesis $h^*$ on both source and target domains. 
Figure \ref{fig:In-depth} (a) and (b) visualize the values of $\mathcal{H} \Delta \mathcal{H}$-distance (measuring domain shift as the discrepancy between the disagreement of two hypotheses $h, h' \in \mathcal{H} \Delta \mathcal{H}$) 
and $\lambda$ (the ideal joint hypothesis $h^*$ on both source and target domains), 
and demonstrate GALA leads to a lower $\mathcal{H} \Delta \mathcal{H}$-distance as well as a lower $\lambda$, indicating a lower generalization error. Note that our GALA balances the classification ability and transferable pattern. 

\section{Conclusions}
\label{sec:conc}


In this paper, we introduce a novel active domain adaptation setting, MS-ADA, which presents a more practical and challenging task by requiring only a few annotated target-domain images to address domain shifts across multiple source domains. To tackle this scenario, we propose the \underline{G}lob\underline{A}l-\underline{L}oc\underline{A}l approach (GALA) to effectively address two key challenges: the performance imbalance caused by variations across source domains, and the style and texture differences between them. Specifically, GALA ensures inter-class diversity in the target domain by focusing on the class distribution within the target dataset and enhances transferability by employing an appropriate target sample selection strategy. The proposed GALA is a simple yet powerful tool that integrates a global $k$-means clustering step for the target domain, followed by a cluster-wise local selection criterion. Importantly, GALA is a plug-and-play tool that can be easily incorporated into existing DA approaches without adding any trainable parameters. Extensive experiments on three benchmark DA datasets demonstrate that GALA outperforms state-of-the-art methods with significant improvements, achieving performance comparable to the fully-supervised upperbound with only 1\% annotation of the target domain, highlighting its efficiency and effectiveness.

\bibliographystyle{IEEEbib}
\bibliography{ref}

\vfill

\end{document}